\documentclass[letterpaper, 10pt, conference]{ieeeconf}

\usepackage[utf8]{inputenc}
\usepackage[T1]{fontenc}
\usepackage{acro}
\usepackage{amsfonts}
\usepackage{amsmath}
\usepackage[keeplastbox, nocheckfootnote]{flushend}
\usepackage{pgfplots}
\usepackage{pgfplotstable}
\usepackage[separate-uncertainty=true]{siunitx}
\usepackage{subcaption}
\usepackage{tikz}
\usepackage{url}

\usetikzlibrary{calc, shapes.misc, arrows, positioning, 3d, scopes}
\pgfplotsset{compat=1.15}
\newcommand*\circled[1]{\tikz[baseline=(char.base)]{\node[shape=circle, draw, inner sep=2pt] (char) {#1};}}

\DeclareAcronym{FCNN}{short=FCNN, long=fully-convolutional neural network}
\DeclareAcronym{GAN}{short=GAN, long=generative adversarial network}
\DeclareAcronym{MDP}{short=MDP, long=Markov decision process}
\DeclareAcronym{NN}{short=NN, long=neural network}
\DeclareAcronym{RL}{short=RL, long=reinforcement learning}
\DeclareAcronym{PPH}{short=PPH, long=picks per hour}
\DeclareAcronym{TCP}{short=TCP, long=tool center point}
\DeclareAcronym{VAE}{short=VAE, long=variational autoencoder}

\DeclareMathOperator*{\argmax}{arg\,max}

\title{\LARGE \bf
Learning a Generative Transition Model \\ for Uncertainty-Aware Robotic Manipulation
}

\author{
    Lars Berscheid$^{1}$, Pascal Meißner$^{2}$, and Torsten Kröger$^{1}$
    \thanks{
        $^{1}$Karlsruhe Institute of Technology (KIT)
	    {\tt\small \{lars.berscheid, torsten\}@kit.edu}}
	\thanks{$^{2}$University of Aberdeen
	    {\tt\small pascal.meissner@abdn.ac.uk}
	}
}

\IEEEoverridecommandlockouts

\begin{document}

\maketitle

\begin{abstract}
Robot learning of real-world manipulation tasks remains challenging and time consuming, even though actions are often simplified by single-step manipulation primitives. In order to compensate the removed time dependency, we additionally learn an image-to-image transition model that is able to predict a next state including its uncertainty. We apply this approach to bin picking, the task of emptying a bin using grasping as well as pre-grasping manipulation as fast as possible. The transition model is trained with up to \num{42000} pairs of real-world images before and after a manipulation action. Our approach enables two important skills: First, for applications with flange-mounted cameras, picks per hours (PPH) can be increased by around \SI{15}{\%} by skipping image measurements. Second, we use the model to plan action sequences ahead of time and optimize time-dependent rewards, e.g.\ to minimize the number of actions required to empty the bin. We evaluate both improvements with real-robot experiments and achieve over \SI{700}{PPH} in the YCB Box and Blocks Test.
\end{abstract}


\section{INTRODUCTION}

For real-world manipulation tasks, a robot needs to deal with unknown, stochastic, and contact-rich environments as well as occluded and noisy sensor data. To approach these challenging constraints, a common simplification is to omit time dependency by decomposing actions into single-step, open-loop manipulation primitives. While this is done for classical analytical approaches, it is even more important for learned robotic manipulation. Recent innovations allow robots to learn tasks by interacting with its environment and maximizing the obtained reward. Since data consumption is the fundamental limitation in most learning tasks, the task itself needs to be as simple as possible. Single-step primitives allow to reduce costly training, in particular in the real world.

However, the discarded time dependency is often useful for practical applications. Amongst others, it first allows to plan longer manipulation sequences or second, optimize for time-dependent multiple-step criteria. In this work, we propose to keep learning the manipulation primitives in a single-step and data-efficient manner, while introducing a transition model on top. This visual transition model is learned from pairs of images \textit{before} and \textit{after} an executed manipulation action, and is then able to \textit{predict} the resulting state of an action. As a transition model adds errors into the system, we further predict the uncertainty of the transition model. By estimating the final uncertainty of the manipulation action, the robot is able to manipulate in a \textit{risk-aware} manner.

In this work, we address the task of \textit{bin picking} using a flange-mounted depth camera. It highlights several challenges of robotic grasping, e.g. unknown and partially hidden objects, as well as an obstacle-rich environment. On the basis of our prior work, we learn both the task of grasping as well as simple pre-grasping manipulation like shifting or pushing in real-world experiments \cite{berscheid_shifting_2019}.

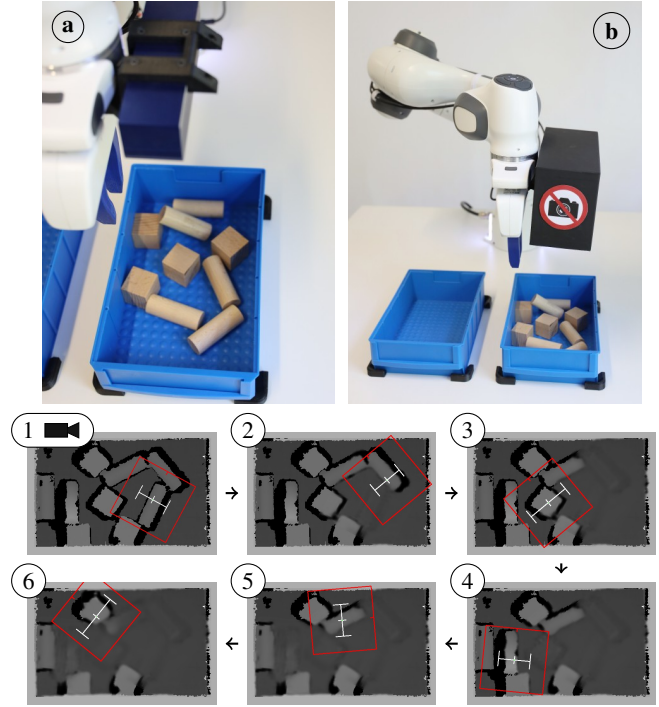
\begin{figure}[t]
    \centering

\newcommand\IncludeGraphic[1]{
    \includegraphics[trim=88 42 65 50, clip, width=0.29\linewidth]{figures/simple-example-large-1/#1}
}

\begin{tikzpicture}
    \node (c) {};
    
	\node[inner sep=0, left of=c, node distance=58] (front-left) {\includegraphics[trim=10 56 26 54, clip, width=0.45\linewidth]{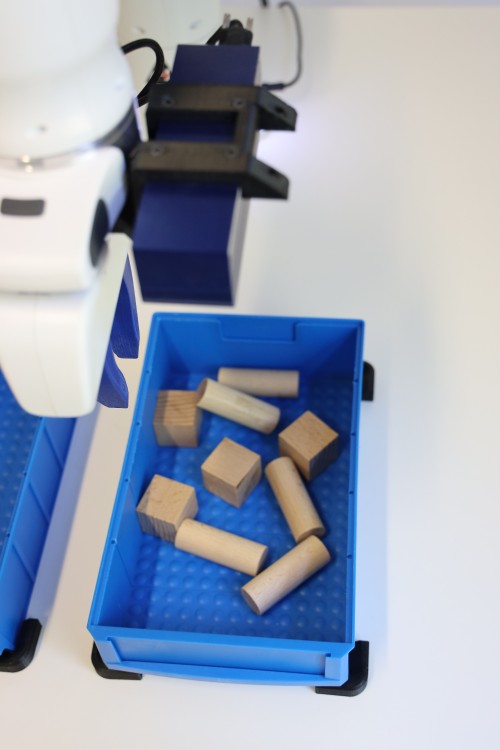}};
	
	\node[inner sep=0, right of=c, node distance=58] (front-right) {\includegraphics[trim=10 56 26 54, clip, width=0.45\linewidth]{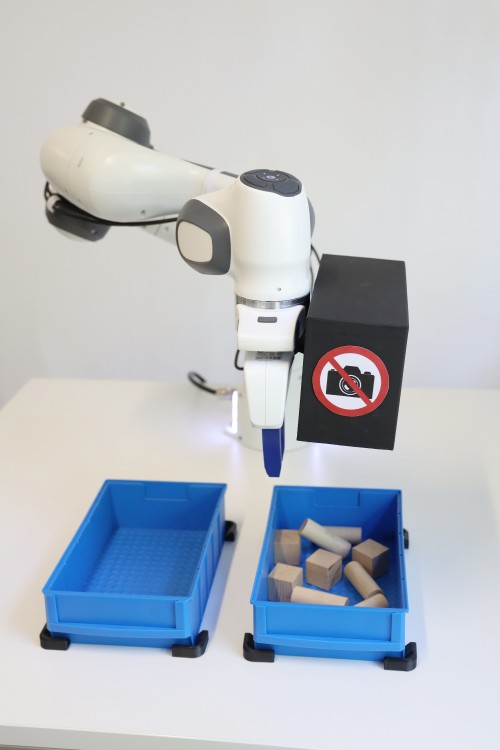}};
	
	\node[below of=c, node distance=110] (2)                {\IncludeGraphic{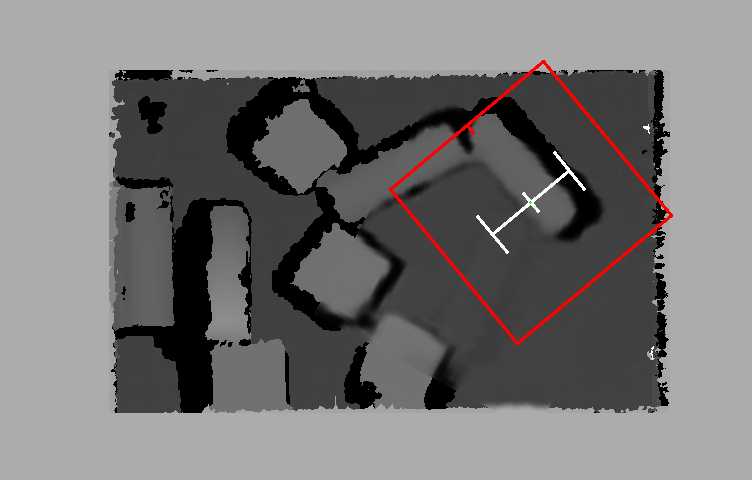}};
	\node[left of=2, node distance=0.338\linewidth] (1)      {\IncludeGraphic{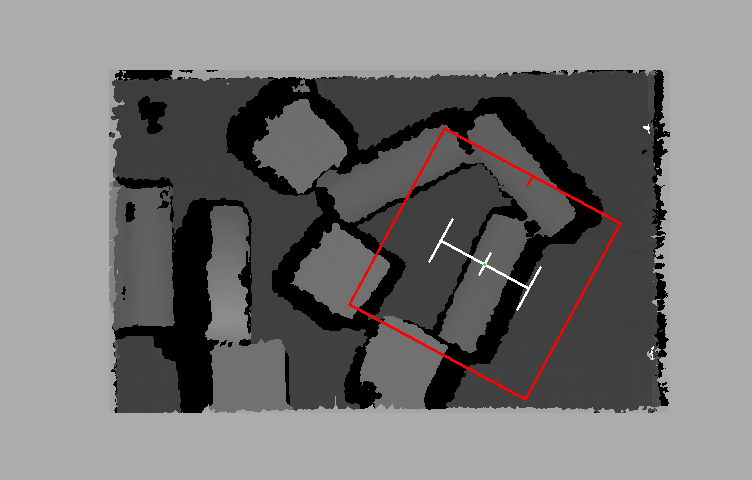}};
	\node[right of=2, node distance=0.338\linewidth] (3)     {\IncludeGraphic{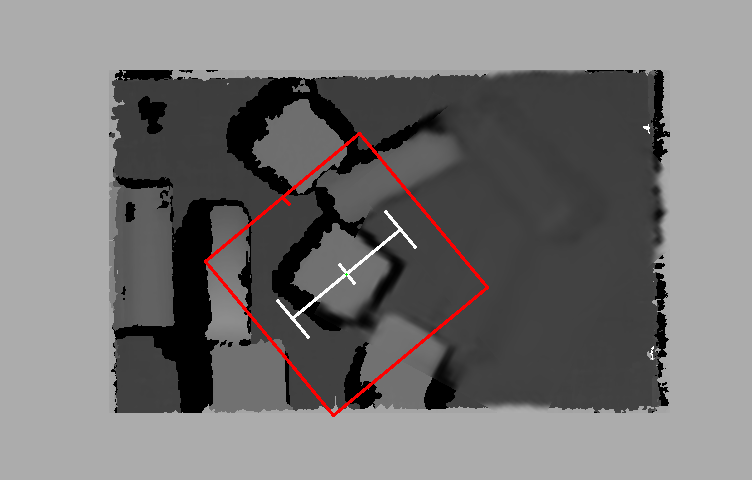}};
	
	\node[] at ($(front-left.north west) + (0.35, -0.35)$) {\small \circled{\textbf{a}}};
	\node[] at ($(front-right.north east) + (-0.4, -0.4)$) {\small \circled{\textbf{b}}};

	\node[below of=3, node distance=57] (4) {\IncludeGraphic{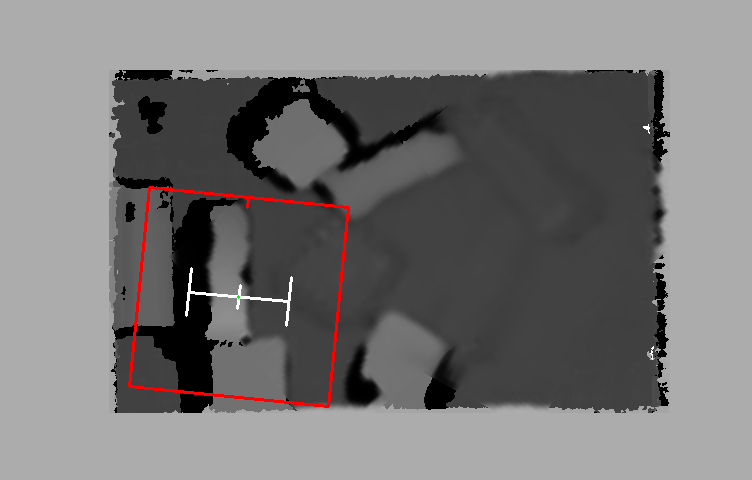}};
	\node[below of=2, node distance=57] (5) {\IncludeGraphic{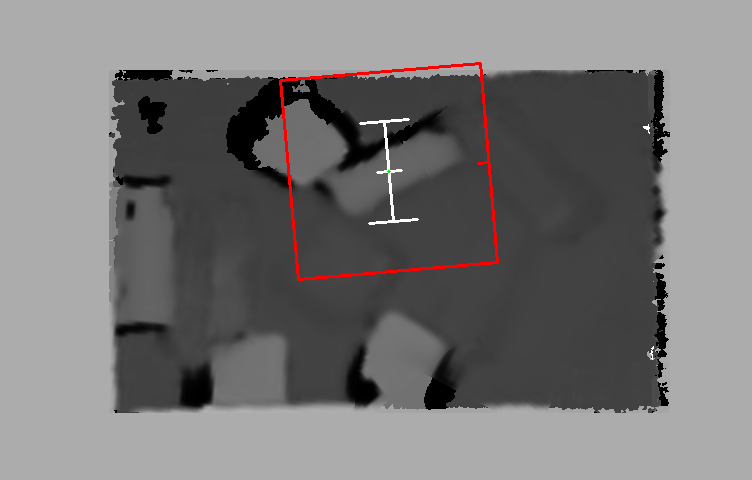}};
	\node[below of=1, node distance=57] (6) {\IncludeGraphic{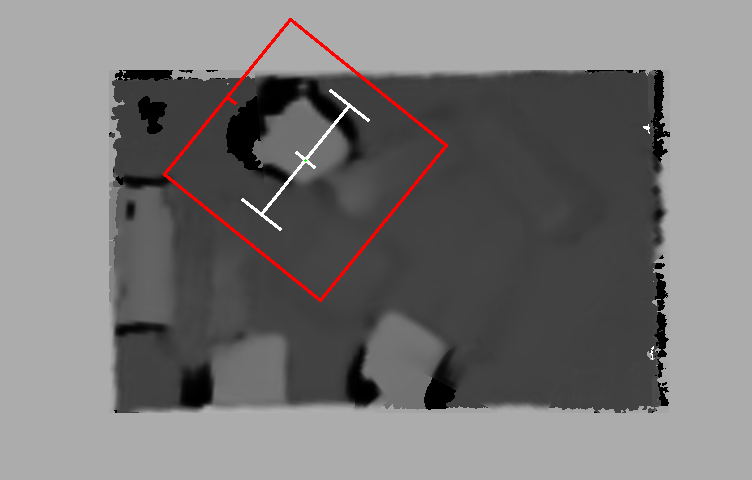}};
	
	\coordinate (diff) at (-0.1, -0.1);
	\tikzstyle{number} = [rounded rectangle, draw=black, fill=white, anchor=west]
	\node[number, text width=0.8cm] (cam) at ($(1.north west) + (diff)$) {\small 1};
	\node[number] at ($(2.north west) + (diff)$) {\small 2};
	\node[number] at ($(3.north west) + (diff)$) {\small 3};
	\node[number] at ($(4.north west) + (diff)$) {\small 4};
	\node[number] at ($(5.north west) + (diff)$) {\small 5};
	\node[number] at ($(6.north west) + (diff)$) {\small 6};
	
	\draw [thick, ->] (1.east) to (2.west);
	\draw [thick, ->] (2.east) to (3.west);
	\draw [thick, ->] (3.south) to (4.north);
	\draw [thick, ->] (4.west) to (5.east);
	\draw [thick, ->] (5.west) to (6.east);
	
	\draw [fill=black!92] ($(cam.west) + (0.45, -0.09)$) rectangle +(0.3, 0.18);
	\draw [fill=black!92] ($(cam.west) + (0.75, 0)$) ++(0, 0.03) -- ++(0.15, 0.07) -- ++(0, -0.2) -- ++(-0.15, 0.07);
\end{tikzpicture}

	\caption[]{After a single measurement (a), the robot predicts the next visual state using a learned generative transition model (1-6) and is able to grasp objects blindly considering the predicted uncertainty (b). Image (1) was taken by the flange-mounted depth camera and shows the chosen grasps (white) and the relevant input window for the algorithm (red). \footnotemark[2]}
	\label{fig:front-page}
\end{figure}

We see our contributions as follows: First, we introduce a transition model for predicting both the visual state \textit{and} its pixel-wise uncertainty after a manipulation action. Second, we propagate this uncertainty through the manipulation model. For the latter, we use a common fully-convolutional reward estimation for planar manipulation. The robot is then able to plan actions on the predicted state regarding their uncertainty. Third, we evaluate the novel abilities in various real-world robotic experiments, e.g. by increasing the \ac{PPH} in bin picking tasks.

\footnotetext[2]{Supplementary material is published at \url{https://pantor.github.io/learning-transition-for-manipulation}}

\section{RELATED WORK}

Grasping as a foundation for robotic manipulation has been of great interest since the beginning of robotics research. Bohg et al.~\cite{bohg_data-driven_2014} differentiate between analytical and data-driven approaches. Historically, grasps were synthesized commonly based on analytical constructions of force-closure. For known objects, sampling and ranking based on model-based grasp metrics is still widespread \cite{miller_graspit!_2004}. However, as modeling grasps itself is challenging, it is even harder to do so for pre-grasping manipulation. Moreover, we will focus on the case of manipulation without object model.

\textbf{Learning for manipulation}: In recent years, learning as a purely data-driven method has achieved great progress for manipulation. Kroemer et al.~\cite{kroemer2019review} gave a survey about the variety of existing approaches. From our perspective, two major trends have emerged: First, an \textit{end-to-end} approach using a step-wise, velocity-like control of the end effector \cite{levine_learning_2016, jang_end--end_2017, kalashnikov_2018_qt, quillen_deep_2018}. Second, the usage of predefined single-step \textit{manipulation primitives} promises a less powerful but more data-efficient learning. This is often combined with planar manipulation and a \ac{FCNN} as a grasp quality \cite{mahler_dex-net_2017} or (more general) an action-value estimator \cite{berscheid_shifting_2019, zeng_learning_2018}. Similar to \textit{Dex-Net} \cite{mahler_dex-net_2017} for bin picking, these approaches map a single image to a single grasp point. Based on the data source, learning for manipulation can be grouped using simulation \cite{quillen_deep_2018}, analytical metrics \cite{mahler_dex-net_2017}, or real-world interaction \cite{berscheid_shifting_2019, zeng_learning_2018}. In particular for real-world learning without any transfer gap, data efficiency often limits manipulation to single-step primitives in practice \cite{berscheid_improving_2019}.

\textbf{Transition model}: In this work, we focus on the \textit{transition model} and its applications for robotic manipulation. Here, model-based \ac{RL} promises data-efficiency by incorporating a learned transition model into the policy training. While model-based \ac{RL} have successfully been applied to either non-robotic tasks \cite{atkeson1997comparison, kaiser2019model, hafner2019learning} or low-dimensional non-visual robotic tasks \cite{abbeel_2007_application}, the high-dimensional visual state space hinders adaption for robotic manipulation. Nevertheless, some ideas were explored in this domain: Byravan et al.~\cite{byravan_2017_se3} learned a model predicting rigid body motions from depth images, enabling to generate further rendered images from an updated state model. Boots et al.~\cite{boots_2014_learning} learned an image-to-image transition model of a moving robot arm. More recently, Finn et al.~\cite{finn2019foresight} learned a model estimating pixel flow transformations on images. Combined with model-predictive control, the robot performed nonprehensile manipulation with unknown objects. Similarly, Ebert et al.~\cite{ebert2018visual} learned a visual transition model to infer actions to reach an image-specified goal for manipulation tasks. 

\textbf{Image-to-image translation}: The core task of a visual transition model is image-to-image translation. In recent years, two major approaches emerged in the field of machine learning: First, \ac{GAN} based methods allow to generate data samples similar to a training set \cite{goodfellow_generative_2014}. On top, the \textit{Pix2Pix} architecture by Isola et al.~\cite{isola_image-to-image_2016} was used in a variety of applications. It is based on a conditional \ac{GAN} combined with a U-Net architecture for the generating \ac{NN}. However, it fails to capture stochastic output distributions, as it is a deterministic mapping and very sensitive to mode collapse \cite{isola_image-to-image_2016}. Second, \ac{VAE} based methods try to capture the complete data distribution via a probabilistic latent space. Combining \acp{GAN} and \acp{VAE}, Zhu et al.~\cite{zhu_multimodal_2017} introduced the \textit{BicycleGAN} that enables multi-modal output distributions and uncertainty estimation. Furthermore, we'll give a brief introduction, but focus on the integration of the BicycleGAN architecture within our robotics application.

\section{LEARNING A TRANSITION MODEL}

\begin{figure}[t]
    \centering
    \vspace{2mm}
    
    \begin{subfigure}[t]{\linewidth}
    \centering
\begin{tikzpicture}[font=\small, scale=0.75]
    \newcommand{\layer}[6][] {
		\draw[#1] (#2,#3 / 2,#4 / 2) -- ++(#5,0,0) -- ++(0, 0, -#4) -- ++(-#5, 0, 0) -- cycle;
		\draw[#1] (#2,#3 / 2,#4 / 2) -- ++(#5,0,0) -- ++(0, -#3, 0) -- ++(-#5, 0, 0) -- cycle;
		\draw[#1] (#2 + #5,-#3 / 2,-#4 / 2) -- ++(0,0,#4) -- ++(0,#3,0) -- ++(0,0,-#4) -- cycle;
	}
	
	\newcommand{\window}[7][] {
		\draw[#1] (#2 + #5,-#3 / 2 + #6,-#4 / 2 + #7) -- ++(0,0,#4) -- ++(0,#3,0) -- ++(0,0,-#4) -- cycle;
	}
	
	\layer[fill=white, rotate around x=-75]{-1.7}{3.6}{3.6}{0}{}
	\layer[fill=white, rotate around x=-50]{-1.3}{3.6}{3.6}{0}{}
	\layer[fill=white, rotate around x=-25]{-0.9}{3.6}{3.6}{0}{}
	\node[] () at (-1.2, -1.5) {$\cdots$};
	
	\layer[thick, fill=white]{0}{3.6}{3.6}{0.07}{}
	\begin{scope}[canvas is yz plane at x=0.07]
		\node[transform shape, rotate=-90] (a) {\includegraphics[width=3.6cm, height=3.6cm]{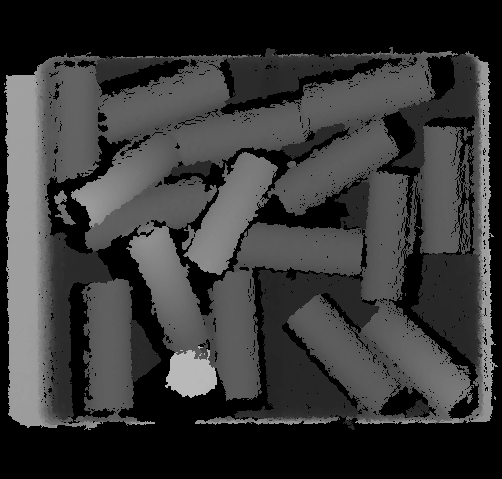}};
	\end{scope}
	

    \draw[blue, thick, fill=white] (1.4, -1.2) rectangle (1.6, 1.2);
    \draw[blue, thick, fill=white] (1.9, -0.7) rectangle (2.1, 0.7);
    \draw[blue, thick, fill=white] (2.4, -0.4) rectangle (2.7, 0.4);
    \draw[blue, thick, fill=white] (3.0, -0.2) rectangle (3.4, 0.2);
    \draw[blue, thick, fill=white] (3.7, -0.1) rectangle (4.5, 0.1);
    
    \draw[thick, ->] (0.8, 0) -- (1.4, 0);
    \draw[thick, ->] (1.6, 0) -- (1.9, 0);
    \draw[thick, ->] (2.1, 0) -- (2.4, 0);
    \draw[thick, ->] (2.7, 0) -- (3.0, 0);
    \draw[thick, ->] (3.4, 0) -- (3.7, 0);
    \draw[thick, ->] (4.5, 0) -- (5.0, 0);
	
	\layer[thick, fill=white]{5.63}{3.6}{3.6}{0.07}{}
	\begin{scope}[canvas is yz plane at x=5.7]
		\node[transform shape, rotate=-90] (a) {\includegraphics[width=3.6cm, height=3.6cm]{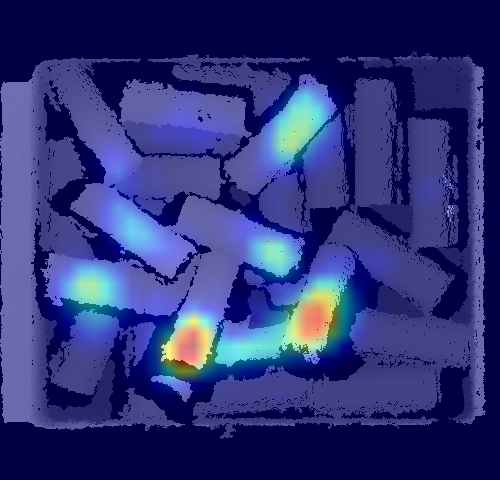}};
	\end{scope}
\end{tikzpicture}
    \caption{Manipulation Model $M$: The fully-convolutional manipulation \acf{NN} efficiently estimates the reward for a large number of possible actions. For a single rotation and single primitive type, the result can be interpreted as a reward heatmap.}
    \label{fig:manipulation-model-arch}
    \end{subfigure}
~
    \begin{subfigure}[t]{\linewidth}
    \centering
\begin{tikzpicture}[font=\small, scale=0.75]
    \newcommand{\layer}[6][] {
		\draw[#1] (#2,#3 / 2,#4 / 2) -- ++(#5,0,0) -- ++(0, 0, -#4) -- ++(-#5, 0, 0) -- cycle;
		\draw[#1] (#2,#3 / 2,#4 / 2) -- ++(#5,0,0) -- ++(0, -#3, 0) -- ++(-#5, 0, 0) -- cycle;
		\draw[#1] (#2 + #5,-#3 / 2,-#4 / 2) -- ++(0,0,#4) -- ++(0,#3,0) -- ++(0,0,-#4) -- cycle;
	}
	
	\newcommand{\window}[7][] {
		\draw[#1] (#2 + #5,-#3 / 2 + #6,-#4 / 2 + #7) -- ++(0,0,#4) -- ++(0,#3,0) -- ++(0,0,-#4) -- cycle;
	}
	
	\layer[thick, fill=white]{0}{3.6}{3.6}{0.07}{}
	\begin{scope}[canvas is yz plane at x=0.077]
		\node[transform shape, rotate=-90] (a) {\includegraphics[width=3.6cm, height=3.6cm]{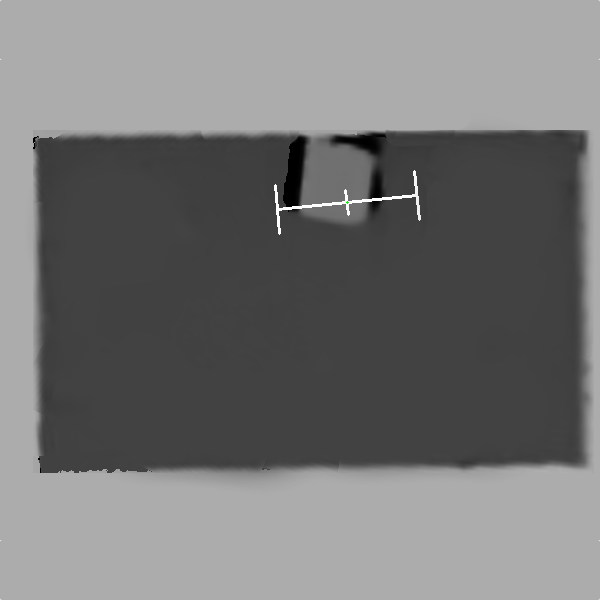}};
	\end{scope}
	
	\window[red, thin]{0}{1.4}{1.4}{0.08}{0.55}{-0.3}

    \draw[blue, thick, fill=white] (1.5, -1.2) rectangle (1.8, 1.2);
    \draw[blue, thick, fill=white] (2.1, -0.5) rectangle (2.4, 0.5);
    \draw[blue, thick, fill=white] (2.7, -0.2) rectangle (3.3, 0.2);
    \draw[blue, thick, fill=gray!20] (3.52, -0.6) rectangle (3.82, 0.4);
    \draw[blue, thick, fill=white] (3.6, -0.5) rectangle (3.9, 0.5);
    \draw[blue, thick, fill=gray!20] (4.12, -1.3) rectangle (4.42, 1.1);
    \draw[blue, thick, fill=white] (4.2, -1.2) rectangle (4.5, 1.2);
    
    \draw[thick, ->] (0.8, 0) -- (1.5, 0);
    \draw[thick, ->] (1.8, 0) -- (2.1, 0);
    \draw[thick, ->] (2.4, 0) -- (2.7, 0);
    \draw[thick, ->] (3.3, 0) -- (3.6, 0);
    \draw[thick, ->] (3.9, 0) -- (4.2, 0);
    \draw[thick, ->] (4.5, 0) -- (5.0, 0);
    \draw[thick, ->, dotted] (2.4, -0.35) -- (3.6, -0.35);
    \draw[thick, ->, dotted] (1.8, -1.0) -- (4.2, -1.0);
	
	\layer[thick, fill=white]{6.0}{3.6}{3.6}{0.07}{}
	\begin{scope}[canvas is yz plane at x=6.07]
		\node[transform shape, rotate=-90] (a) {\includegraphics[width=3.6cm, height=3.6cm]{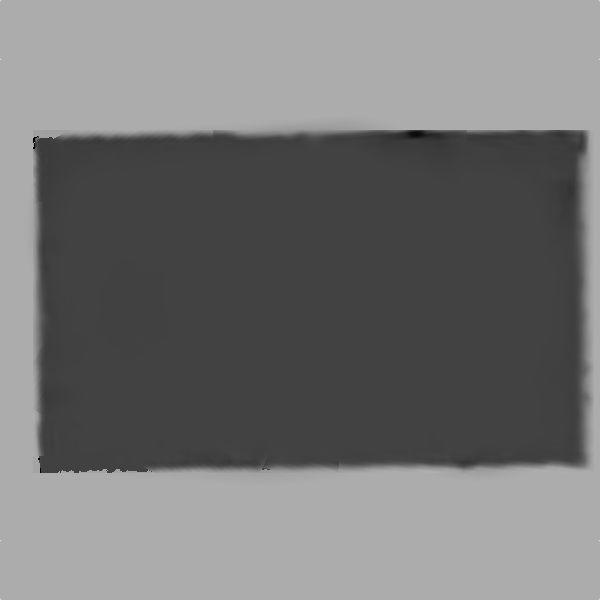}};
	\end{scope}
	\window[red, thin]{6.0}{1.4}{1.4}{0.08}{0.55}{-0.3}
\end{tikzpicture}
    \caption{Transition Model $T$: The \textit{U-Net} architecture predicts the next image state $s_{t+1}$ depending on the prior state $s_{t}$, the action $a_t$, and its reward $r_t$. The result of the predicted window around the action (red) is then patched into the original image.}
    \label{fig:transition-model-arch}
    \end{subfigure}

    \caption{We make use of two \acp{NN}, one for the \textit{manipulation} policy (a) and one for the \textit{transition} model (b).}
    \label{fig:smy_label}
\end{figure}
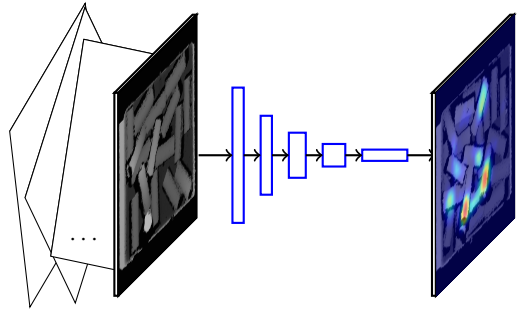
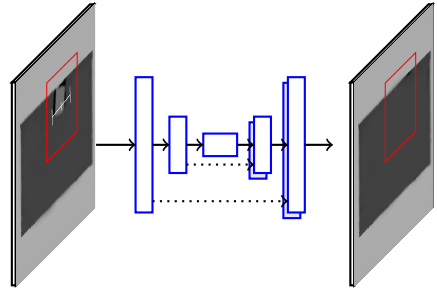

From \Acf{RL}, we adopt the concept of a \ac{MDP} $(\mathcal{S}, \mathcal{A}, T^*, r, p_0)$ with the state space $\mathcal{S}$, the action space $\mathcal{A}$, the transition distribution $T^*$, the reward function $r$ and the initial configuration $p_0$. Similar to other data-driven approaches, \ac{RL} is limited by its data consumption, and even more so for time-dependent tasks resulting in \textit{sparse rewards}. For this reason, we simplify manipulation to a single time step problem. Then, a solution to this \ac{MDP} is a policy $\pi: \mathcal{S} \mapsto \mathcal{A}$ mapping the current state $s \in \mathcal{S}$ to an action $a \in \mathcal{A}$. In this work, we explicitly learn a transition model $T: \mathcal{S} \times \mathcal{A} \mapsto \mathcal{S}$ approximating the true transition distribution $T^*$.

\subsection{Learning for Robotic Manipulation}

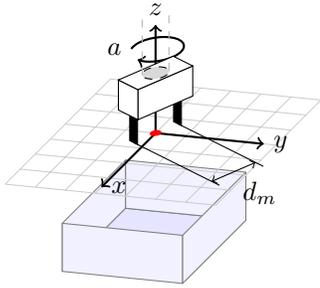
\begin{figure}[t]
	\centering
	\vspace{0.8mm}

	\centering
\begin{tikzpicture}[x={(-0.5cm, -0.5cm)}, y={(1cm, -0.1cm)}, z={(0cm, 1cm)}, scale=0.8]
	\definecolor{bincolor}{RGB}{80, 80, 255}
	\tikzstyle{facestyle} = [fill=bincolor!8, draw=gray, thin, line join=round]
	\tikzstyle{ann} = [black, inner sep=1pt]
	\tikzstyle{coordaxis} = [black, thick, ->]
	
	\begin{scope}[canvas is yx plane at z=0.2]
		\path[facestyle, fill=bincolor!15] (-0.5,0) rectangle (1.5,2);
	\end{scope}
	
	\begin{scope}[canvas is zy plane at x=0]
		\path[facestyle] (0.2,-0.5) rectangle (1,1.5);
	\end{scope}
	
	\begin{scope}[canvas is zx plane at y=-0.5]
		\draw[facestyle] (0.2,0) rectangle (1,2.1);
	\end{scope}
	
	\begin{scope}[canvas is zy plane at x=2.1]
		\path[facestyle] (0.2,-0.5) rectangle (1,1.5);
	\end{scope}
	
	\begin{scope}[canvas is zx plane at y=1.5]
		\draw[facestyle] (0.2,0) rectangle (1,2.1);
	\end{scope}
	
	\begin{scope}[canvas is xy plane at z=1.5]
		\draw[step=5mm, very thin, gray!40] (-1.5,-1.5) grid (2,2);
	\end{scope}
	
	\draw[coordaxis] (0,0,1.5) -- +(1.8,0,0) node[right] {$x$};
	\draw[coordaxis] (0,0,1.5) -- +(0,1.8,0) node[right] {$y$};
    \draw[coordaxis] (0,0,1.5) -- +(0,0,1.8) node[above] {$z$};
    
    \draw[coordaxis, <-] (0.2,-0.2,2.8) arc(-20:280:.4) node[left] {$a$};

	\begin{scope}[shift={(0, 0, 2.0)}, rotate around z=-28]
		\draw[black, fill=black] (0.4,-0.1,0) -- ++(0,0,-0.5) -- ++(0,0.2,0) -- ++(0,0,0.5) -- cycle;
		\draw[black, fill=black] (-0.4,-0.1,0) -- ++(0,0,-0.5) -- ++(0,0.2,0) -- ++(0,0,0.5) -- cycle;
		
		\draw[black, fill=white] (-0.5,0.25,0) -- ++(1,0,0) -- ++(0,0,0.5) -- ++(-1,0,0) -- cycle;
	    \draw[black, fill=white] (0.5,-0.25,0) -- ++(0,0.5,0) -- ++(0,0,0.5) -- ++(0,-0.5,0) -- cycle;
	    \draw[black, fill=white] (-0.5,-0.25,0.5) -- ++(0,0.5,0) -- ++(1,0,0) -- ++(0,-0.5,0) -- cycle;
	    \draw[black, thin, fill=gray!30, dashed] (0,0,0.5) circle (0.2);
	    \draw[gray!60, thin, dashed] (0.173,-0.1,0.5) -- ++(0,0,1);
	    \draw[gray!60, thin, dashed] (-0.173,0.1,0.5) -- ++(0,0,1);
	    
	    \draw[black, thin] (0.4,0,-0.5) -- +(0,2.2,0);
	    \draw[black, thin] (-0.4,0,-0.5) -- +(0,2.2,0);
	    \draw[black, thin, <->] (-0.4,2,-0.5) -- ++(0.8,0,0) node[below right, midway] {$d_m$};
	    
	    \draw[red, fill=red] (0,0,-0.5) circle (0.08);
	\end{scope}
\end{tikzpicture}
	
	\caption{The manipulation \acf{NN} maps an orthographic depth image to an action type (\textit{grasp} or \textit{shift}) and a planar pose $(x, y, z, a)$ with gripper width $d_m$.}
	\label{fig:task-space}
\end{figure}

Similar to our prior work, we make use of six so-called \textit{manipulation primitives}:
\begin{enumerate}
    \item \textbf{Grasps} with four different pre-shaped gripper widths $d_m$. Given a grasp point, the robot approaches on a trajectory parallel to its gripper. It then closes the gripper, (possibly) lifts an object, and measures the grasp success by its internal force sensors. We define a binary reward $r$ for these primitives equal to the grasp success.
    \item \textbf{Shifts} in two different, horizontal directions (relative to the gripper). The robot approaches a given manipulation point, and moves its closed gripper for a few \si{cm}, leading to a shifting or pushing motion. The reward is defined by the difference of the maximal grasp reward before and after the action within a pre-defined neighborhood of the action. 
\end{enumerate}
We simplify the action space to planar manipulation. Then, each action is described by five parameters: The spatial coordinates $(x, y, z, a)$ (as shown by Fig.~\ref{fig:task-space}) and the manipulation primitive type $p_t$. \\

Let $s_t \in \mathcal{S}$ be an orthographic image of the scene at time step $t$. We align the task-space coordinate system with the image frame by using a top-down view of the scene. This way, orthographic images have a key advantage over other projections: Affine transformations of the image $s$ correspond to planar poses $(x,y,a)$ of the robot.

We assume that manipulation (often) depends only on \textit{local} information. We introduce the cropped window $\hat{s} \in \hat{\mathcal{S}} \subset \mathcal{S}$ at the given affine transformation $(x, y, a)$. Its side length is slightly larger than the largest object size. We use a resolution of $(32 \times 32)$ pixels for the image window. Similar to our prior work \cite{berscheid_shifting_2019, berscheid_improving_2019}, the system is learned in a self-supervised manner to maximize the corresponding reward $r$. Therefore, a manipulation \ac{NN} called $M$ is trained to predict the reward $r$ of the image windows $\hat{s}_t$ with given primitive $p_t$. We make use of a fully-convolutional \ac{NN}: During inference, the same manipulation model $M(s_t)$ estimates rewards $\psi_t$ for a large number of poses efficiently in a sliding-window approach (Fig.~\ref{fig:manipulation-model-arch}). Rewards for different primitives $p_t$ are implemented as multiple channels in the output layer. The height $z$ is calculated trivially by a model-based controller. The final action $a_t$ is selected greedily by the policy $\pi$.

\subsection{Simplifications for the Transition Model}

\begin{figure*}[t]
	\centering
	\vspace{1.6mm}
	
	\definecolor{bis}{RGB}{14, 58, 106}
    \definecolor{bis1}{RGB}{194, 144, 54}
    \definecolor{bisp1}{RGB}{246, 192, 77}
    \definecolor{biN}{RGB}{85, 135, 168}
    \definecolor{biQ}{RGB}{246, 192, 77}
    \definecolor{biz}{RGB}{254, 181, 22}
    \definecolor{biL}{RGB}{224, 230, 243}

\newcommand{\graphicssize}{0.17\linewidth}
\newcommand\IncludeGraphic[1]{%
    \includegraphics[trim=88 42 65 50, clip, width=\graphicssize]{figures/simple-example-1/#1}
}

\begin{tikzpicture}
    \tikzstyle{title} = [text width=\graphicssize, align=left, node distance=50, font=\bfseries]
    \tikzstyle{captiontop} = [text width=\graphicssize, align=left, node distance=36]
    \tikzstyle{captionbottom} = [text width=\graphicssize, align=left, node distance=38]
    \tikzstyle{image} = [node distance=31]

    \node[] (1c) {};
	\node[image, above of=1c] (1r) {\IncludeGraphic{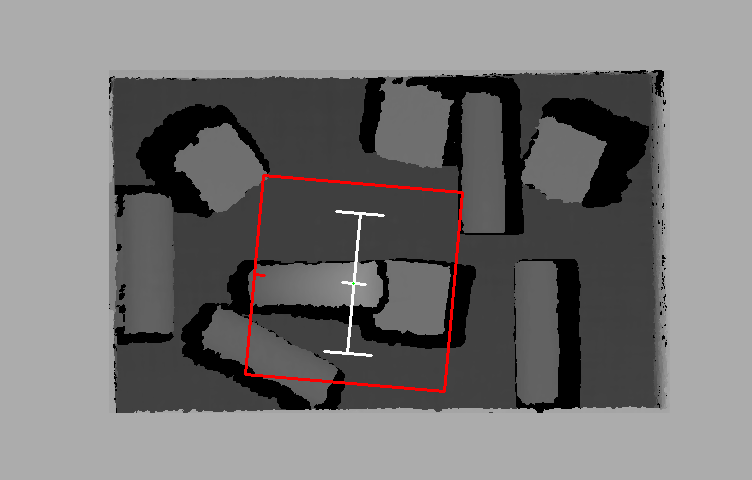}};
	\node[image, below of=1c] (1u) {\IncludeGraphic{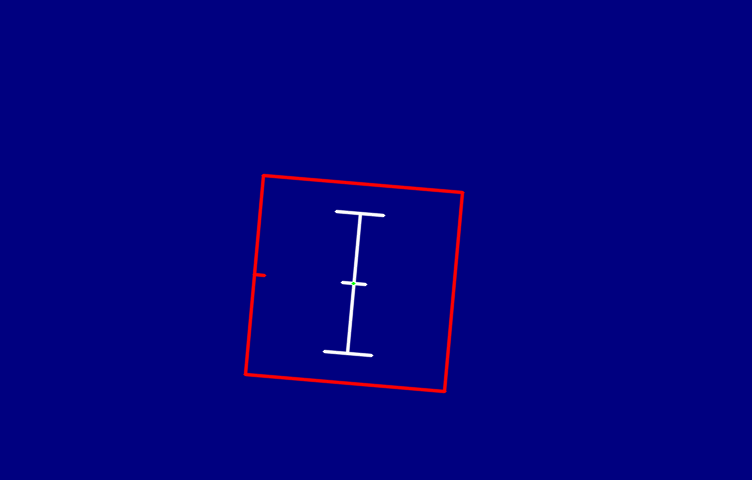}};
	
	\node[captiontop, above of=1r] {\small \textbf{Measurement} $s_{t=0}$};
    \node[captionbottom, below of=1u] {\small Uncertainty $\Sigma_{t=0}^2$};

    \node[draw=black, fill=bis, text=white, minimum width=6mm, minimum height=7mm] (xpsi) at ($(1r.east) + (0.6, -0.3)$) {\small $M$};
    \node[draw=black, fill=bis, text=white, minimum width=6mm, minimum height=7mm] (xpsigrad) at ($(1u.east) + (0.6, 0.3)$) {\small $\frac{\partial M}{\partial s}$};
    
    
    \node[draw=black, fill=biL, right of=1c, node distance=100, minimum width=5mm, minimum height=12mm] (xpi) {\small $\pi$};

    \node[draw=black, fill=bisp1, minimum width=5mm, minimum height=26mm] (xt) at ($(1r.east) + (2.9, -0.5)$) {\small $T$};
    \node[draw=black, fill=biL, rounded rectangle, minimum size=6mm] (xplus) at ($(1u.east) + (3.5, -0.3)$) {\small $+$};

	\node[right of=1c, node distance=214] (2c)  {};
	\node[image, above of=2c] (2r) {\IncludeGraphic{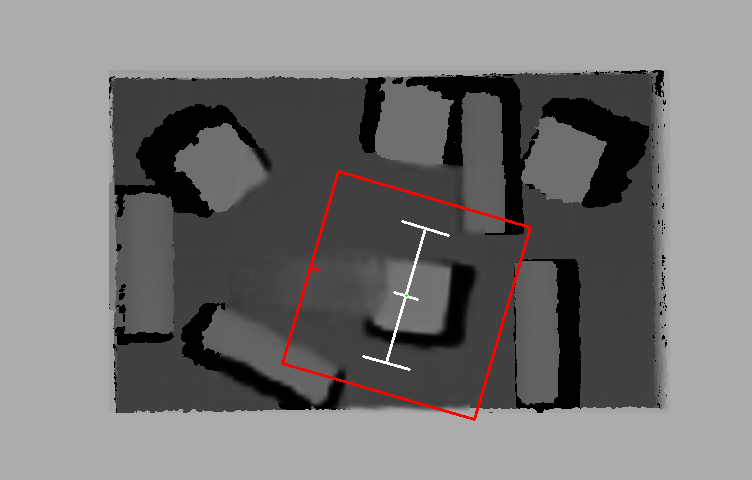}};
	\node[image, below of=2c] (2u) {\IncludeGraphic{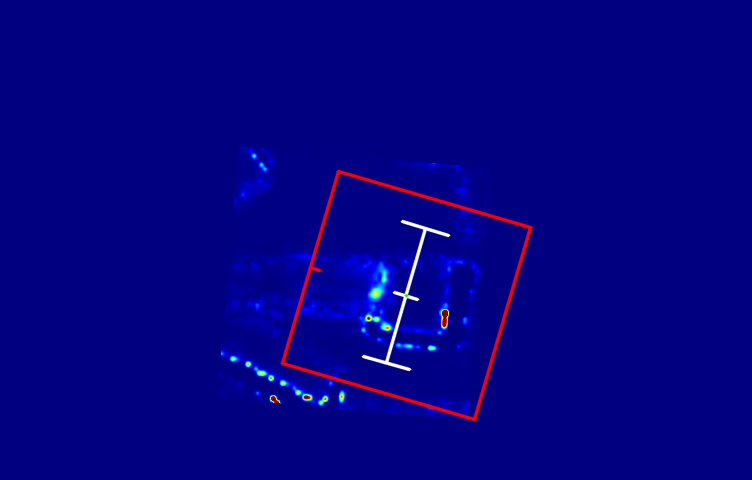}};
	
	\node[captiontop, above of=2r]  {\small \textbf{Prediction} $t=1$};
	
	\node[right of=2c, node distance=100] (3c)  {};
	\node[image, above of=3c] (3r) {\IncludeGraphic{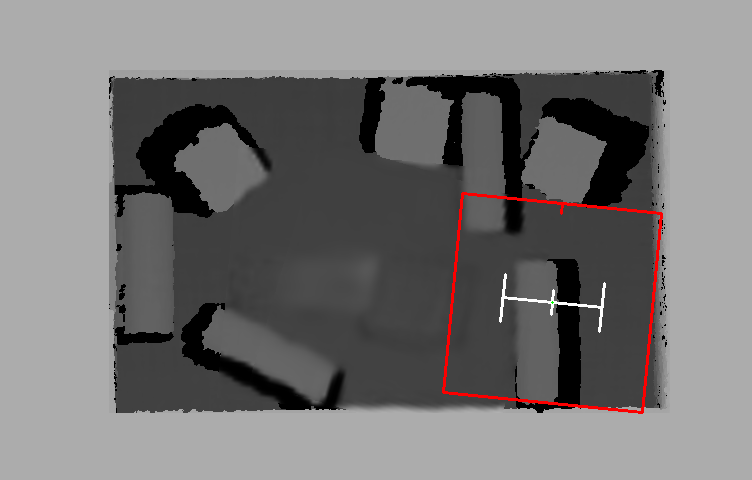}};
	\node[image, below of=3c] (3u) {\IncludeGraphic{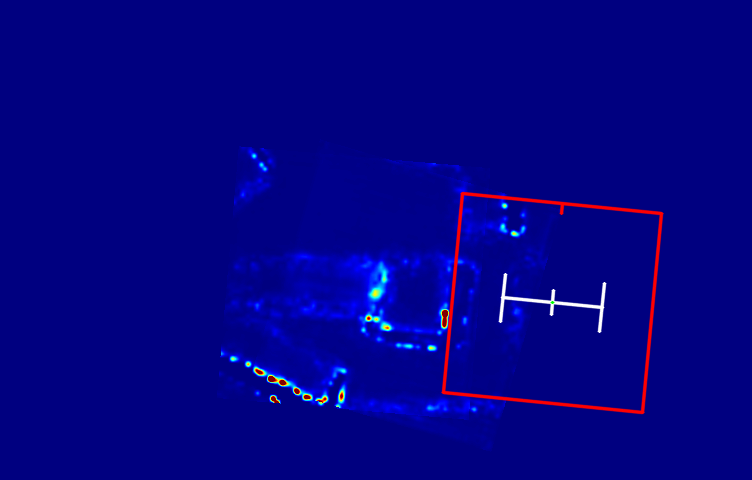}};
	\node[captiontop, above of=3r] {\small $t=2$};
	
	\node[right of=3c, node distance=100] (4c)  {};
	\node[image, above of=4c] (4r) {\IncludeGraphic{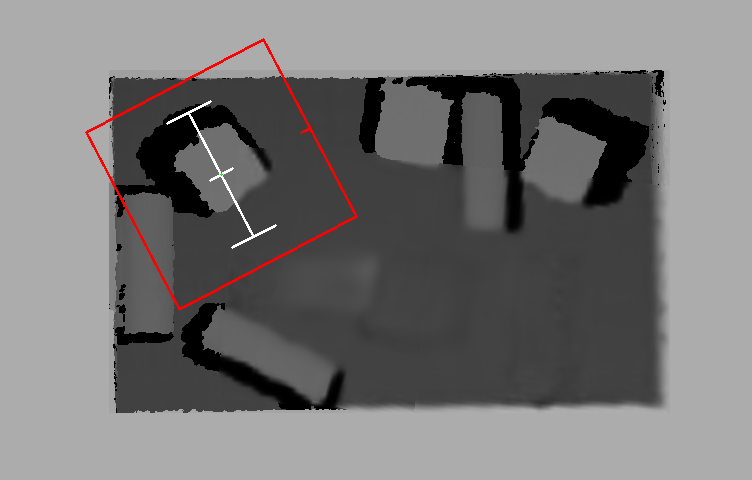}};
	\node[image, below of=4c] (4u) {\IncludeGraphic{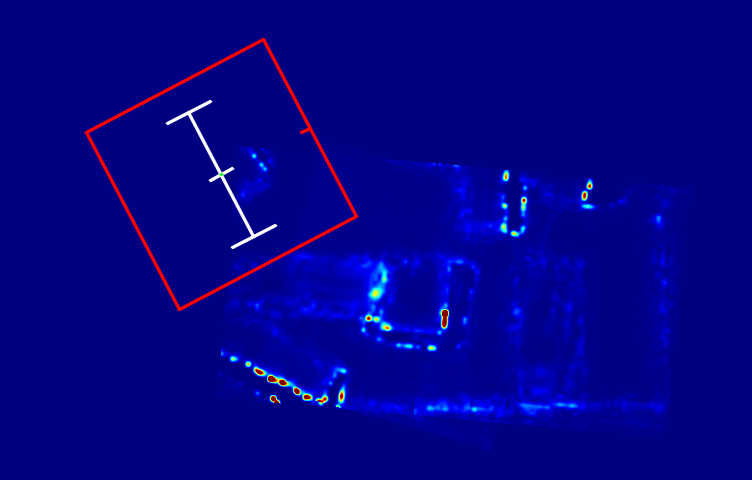}};
	\node[captiontop, above of=4r] {\small $t=3$};

	\begin{scope}[->]
        \draw ([yshift=3mm]1r.east) |- ([yshift=8mm]xt.west);
        \draw ([yshift=-3mm]1r.east) |- (xpsi.west);
        
        \draw ([yshift=3mm]1u.east) |- (xpsigrad.west);
        \draw ([yshift=-3mm]1u.east) |- (xplus.west);
        
        \draw (xpsi.east) -| +(0.2, -0.5) |- ([yshift=3mm]xpi.west);
        \draw (xpsigrad.east) -| +(0.2, 0.5) |- ([yshift=-3mm]xpi.west);
        
        \draw ([yshift=3mm]xpi.east) |- ([yshift=-3mm]xt.west);
        \draw ([yshift=-3mm]xpi.east) |- ([yshift=-9mm]xt.west);
        
        \draw ([yshift=8mm]xt.east) |- ([yshift=3mm]2r.west);
        \draw ([yshift=-9mm]xt.east) -| (xplus.north);
        
        \draw (xplus.east) |- ([yshift=-3mm]2u.west);
    \end{scope}
    
	\begin{scope}[thick, ->, decoration={post length=4pt}, rounded corners=2.5mm]
	    \draw ($(2r.east)!0.5!(2u.east)$) -- ($(3r.west)!0.5!(3u.west)$);
	    \draw ($(3r.east)!0.5!(3u.east)$) -- ($(4r.west)!0.5!(4u.west)$);
    \end{scope}
    
    \node[] at ($(xt.west) + (-0.5, 1.05)$) {$\hat{s}_t$};
    \node[] at ($(xt.east) + (0.45, 1.05)$) {$\hat{s}_{t+1}$};
    \node[] at ($(xt.east) + (0.6, -0.68)$) {$\hat{\sigma}_{s, t+1}^2$};
    
    \node[] at ($(xplus.west) + (-1.0, -0.3)$) {$\hat{\Sigma}_{t}^2$};
    
    \node[] at ($(xpi.east) + (0.26, 0.47)$) {$r_t$};
    \node[] at ($(xpi.east) + (0.26, -0.52)$) {$p_t$};
    
    \node[] at ($(xpsi.east) + (0.44, -0.06)$) {$\psi_t$};
    \node[] at ($(xpsigrad.east) + (0.55, 0.02)$) {$\sigma_{a, t}$};

    \draw[double, densely dashed, ->] ($(4r.north) + (0.7, 0.12)$) -- ++(0.8, 0);
\end{tikzpicture}
	
	\caption{Iterative multiple-step predictions of the state $s_t$ with corresponding uncertainty $\Sigma_t^2$ given an initial measurement $s_{t=0}$. The computational flow during a prediction step is as follows: The manipulation \acf{NN} $M$ estimates rewards $\psi_t$ for a discrete set of action primtives $p_t$, its gradients are used to propagate the cumulative input uncertainty $\Sigma_t^2$ towards the estimated reward $\sigma_{a,t}$. The policy $\pi$ chooses the final action $a$ with corresponding reward $r$ using the lower bound of the estimated reward. Then, the transition model $T$ predicts the new state $\hat{s}_{t+1}$ around the action $a$ with pixel-wise uncertainties $\hat{\sigma}_{s,t+1}^2$. The uncertainties are summed up ranging from low (blue) to high (red).}
	\label{fig:example-iterative-predictions}
\end{figure*}
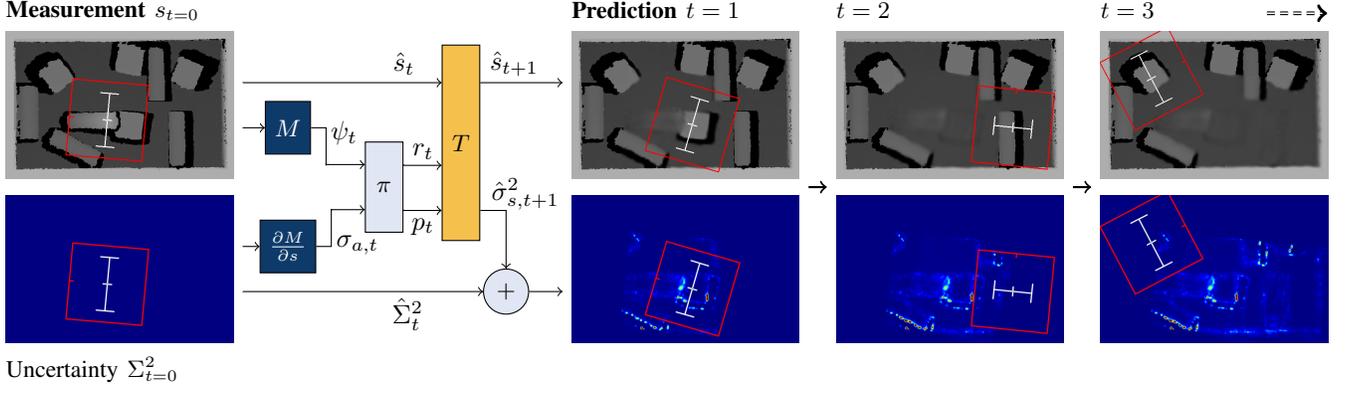

Similar to the local simplifications for manipulation, the transition model $T$ learns only to predict a window $\hat{s}$ around the action pose. While the window size could in principle match to the overall image $s$, we limit the window size to twice the maximal object size. This way, we presume a principle of locality for the state transitions, but it allows to remove all spatial dependencies from the transition model. To estimate the distribution over $s_{t+1}$, the state image $s_{t}$ is patched with $\hat{s}_{t+1}$ at the affine transformation $(x, y, a)$
\begin{align}
	p(s_{t+1}) &= T(s_t, a_t) \approx T(\hat{s}_t, p_t, r_t)
\end{align}
with the manipulation primitive type $p_t$. Additionally, we condition the transition model on the reward $r_t$ of the action. This is motivated by following arguments:
\begin{itemize}
    \item By using the estimated reward $\psi_t$ instead, the calculations of this difficult and high-level feature can be outsourced from the transition model to the already learned manipulation \ac{NN}. This is in particular important for grasp actions: Here, the next states depends strongly on the binary grasp success. Therefore, we train the manipulation model to learn the grasp success explicitly (as it is measurable) and reuse this information in the transition model.
    \item Real reward measurements $r_t$ can be used in the transition model. For example, the grasp reward can be quickly measured by the robot's gripper.
    \item Note that there is no loss of generality by condition the transition model on the reward, as we can substitute the manipulation model as a reward predictor.
\end{itemize}
If not stated otherwise, we simplify the transition model to
\begin{align}
	p(s_{t+1}) &\approx T \left( \hat{s}_t, p_t, M(\hat{s}_t) \right)
\end{align}
Let $\hat{s}_{t+1}^\prime$ denote the prediction with highest probability.

\subsection{Generative Transition Model Architecture}

We employ the BicycleGAN architecture for our transition model \cite{zhu_multimodal_2017}. In a nutshell, the architecture is based on a generator, discriminator, and an encoder \ac{NN}. The generator tries to mimic images from the underlying distribution, taking a noise vector $z$ from a prior latent distribution as input. While playing a min-max-game, the discriminator aims to distinguish the generated data from the real data. The BicycleGAN introduces an encoder to calculate a representation from an image within the latent space $z$. By constraining the latent distribution between the real and generated images, as well as a similarity to a Gaussian, the architecture is able to output a multi-modal distribution. This way, the BicycleGAN architecture is in principle able to capture the stochastic physics as a transition distribution. \\

We use the generator \ac{NN} as the transition model $T$. It uses a \textit{U-Net} architecture with $(64 \times 64)$ pixel input and output size (Fig.~\ref{fig:transition-model-arch}). We extend the BicycleGAN architecture by additionally condition the generator on the manipulation primitive type $p_t$ (as a one-hot encoded vector) and the reward $r_t$.

\subsection{Uncertainty Prediction}

Given the image distribution $p(s_{t+1})$, we are able to estimate a pixel-wise uncertainty of the prediction by sampling from its latent space $z$. Let the uncertainty $\hat{\sigma}_{s}^2$ be the variance of the prediction $\hat{s}_{t+1}^\prime$
\begin{align}
    \hat{\sigma}_{s}^2 &= \mathbb{E} \left[ \left( \hat{s} - \hat{s}_{t+1}^\prime \right)^2 \right] = \sum_i^N p(\hat{s}_i) \left( \hat{s}_{t+1, i} - \hat{s}_{t+1}^\prime \right)^2
\end{align}
given the number of samples $N$. The state uncertainty $\sigma_{s, t+1}^2$ of the overall image is estimated by zero-padding the uncertainty window $\hat{\sigma}_{s}^2$ at $(x, y, a)$. The state uncertainty $\sigma_{s}^2$ can be propagated towards the estimated reward $\psi_t$ of the manipulation action. A first-order Taylor series of the manipulation \ac{NN} $M$ yields
\begin{align}
    \sigma_a^2 \approx \sum_{i, j} \left\lvert \frac{\partial M(\hat{s}_{t+1}^\prime)}{\partial s_{i, j}} \right\rvert^2 \sigma_{s; i, j}^2
    \label{eq:propagation-of-uncertainty}
\end{align}
summing over all pixels $(i, j)$. The approximation presumes a zero covariance between individual pixels and only small uncertainties $\sigma_s^2$. A key contribution of our work is to implement (\ref{eq:propagation-of-uncertainty}) fully-convolutionally: The gradient of the fully-convolutional manipulation \ac{NN} regarding the pixel-wise image input is fully-convolutional itself. Then, the gradient is convoluted over the mean prediction and multiplied with the pixel-wise uncertainty. This results in an efficient computation of both the estimated rewards and its uncertainties $\psi_t \pm \sigma_a$. Note that we only consider the input uncertainty; other uncertainties, in particular from $M$ itself, are neglected in this work.

\subsection{From Predicting to Planning}
\label{subsec:from-predicting-to-planning}

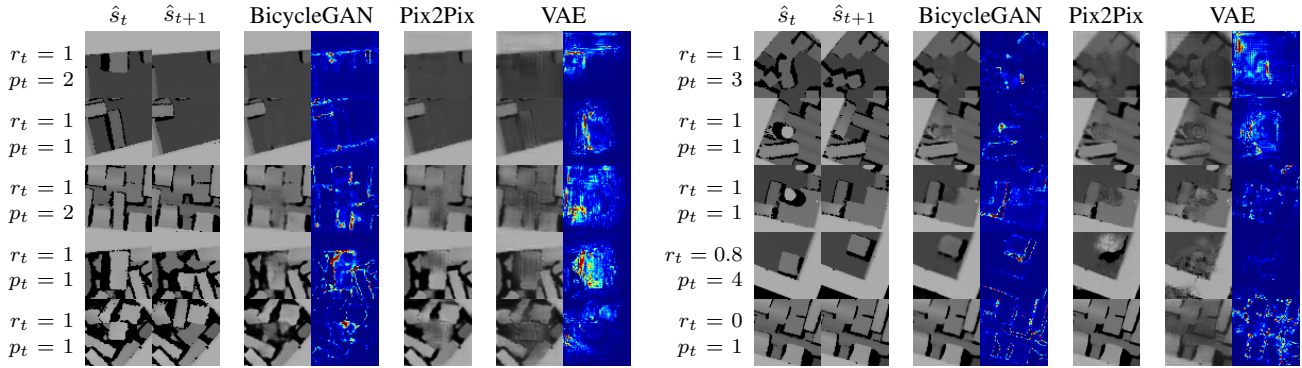
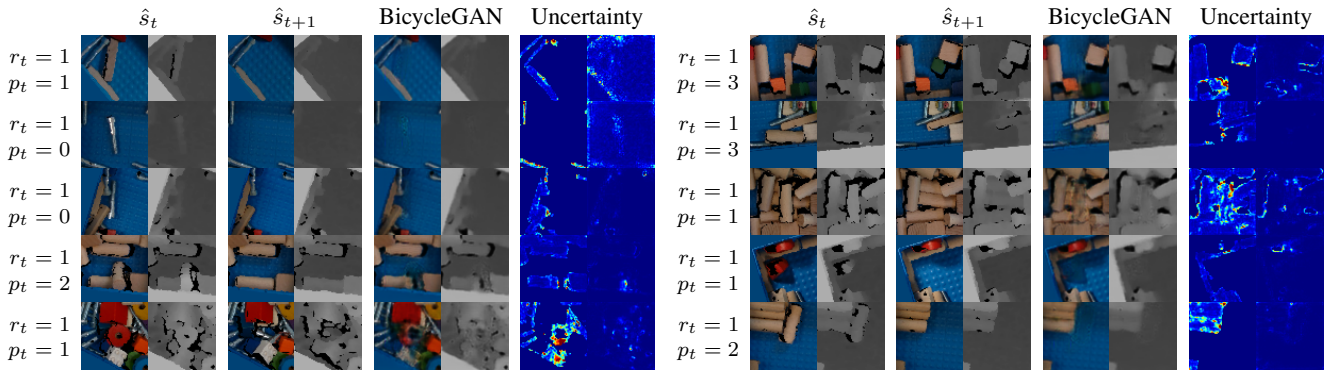
\begin{figure*}
	\centering
	\vspace{0.5mm}

\newcommand{\graphicssize}{0.05\linewidth}
\newcommand\IncludeGraphic[2]{
    \includegraphics[width=\graphicssize]{figures/example-predict/#1/#2}
}

\newcommand\IncludeRGBDGraphic[2]{
    \includegraphics[width=\graphicssize]{figures/example-rgbd/#1/#2}
}

\newcommand\AddRow[5]{
    \node[pred, below of=#3] (#2o) {\IncludeGraphic{#1}{bi_s_bef.png}};
	\node[pred, right of=#2o] (#2a) {\IncludeGraphic{#1}{bi_s_aft.png}};
	
	\node[left of=#2o, node distance=32] (#2tc) {};
	\node[text width=30, align=right, above of=#2tc, node distance=4] {\footnotesize #4};
	\node[text width=30, align=right, below of=#2tc, node distance=6] {\footnotesize #5};

	\node[pred, right of=#2a, node distance=35] (#2b1) {\IncludeGraphic{#1}{bi_result.png}};
	\node[pred, right of=#2b1] (#2bu) {\IncludeGraphic{#1}{bi_unc.png}};
	\node (#2bc) at ($(#2b1)!0.5!(#2bu)$) {};

	\node[pred, right of=#2bu, node distance=35] (#2p1) {\IncludeGraphic{#1}{pix2pix_result.png}};
	\node (#2pc) at ($(#2p1)$) {};
	
	\node[pred, right of=#2p1, node distance=35] (#2v1) {\IncludeGraphic{#1}{vae_result.png}};
	\node[pred, right of=#2v1] (#2vu) {\IncludeGraphic{#1}{vae_unc.png}};
	\node (#2vc) at ($(#2v1)!0.5!(#2vu)$) {};
}

\newcommand\AddColorRow[5]{
    \node[pred, below of=#3] (#2colorbefore) {\IncludeRGBDGraphic{#1}{bi_s_bef_color.png}};
	\node[pred, right of=#2colorbefore] (#2depthbefore) {\IncludeRGBDGraphic{#1}{bi_s_bef_depth.png}};
	\node (#2imagecenter) at ($(#2colorbefore)!0.5!(#2depthbefore)$) {};
	
	\node[left of=#2colorbefore, node distance=29] (#2labelcenter) {};
	\node[text width=24, align=right, above of=#2labelcenter, node distance=4] {\footnotesize #4};
	\node[text width=24, align=right, below of=#2labelcenter, node distance=6] {\footnotesize #5};
	
	\node[pred, right of=#2depthbefore, node distance=30] (#2colorafter) {\IncludeRGBDGraphic{#1}{bi_s_aft_color.png}};
	\node[pred, right of=#2colorafter] (#2depthafter) {\IncludeRGBDGraphic{#1}{bi_s_aft_depth.png}};
	\node (#2colorcenter) at ($(#2colorafter)!0.5!(#2depthafter)$) {};

	\node[pred, right of=#2depthafter, node distance=30] (#2colorpred) {\IncludeRGBDGraphic{#1}{bi_result_color.png}};
	\node[pred, right of=#2colorpred] (#2depthpred) {\IncludeRGBDGraphic{#1}{bi_result_depth.png}};
	\node (#2bc) at ($(#2colorpred)!0.5!(#2depthpred)$) {};
	
	\node[pred, right of=#2depthpred, node distance=30] (#2colorunc) {\IncludeRGBDGraphic{#1}{bi_unc_color.png}};
	\node[pred, right of=#2colorunc] (#2depthunc) {\IncludeRGBDGraphic{#1}{bi_unc_depth.png}};
	\node (#2predcolorcenter) at ($(#2colorunc)!0.5!(#2depthunc)$) {};
}

	\begin{subfigure}[t]{\linewidth}
		\centering
\begin{tikzpicture}
    \tikzstyle{pred} = [node distance=\graphicssize]
    \tikzstyle{textbox} = [node distance=19]
    
    \node[] (cleft) {};

	\AddRow{2019-06-06-13-49-56-259}{1}{cleft}{$r_{t} = 1$}{$p_{t} = 2$}
	\AddRow{2019-06-06-17-55-29-592}{2}{1o}{$r_{t} = 1$}{$p_{t} = 1$}
	\AddRow{2019-07-01-11-09-38-116}{3}{2o}{$r_{t} = 1$}{$p_{t} = 2$}
	\AddRow{2019-03-12-15-17-27-286}{4}{3o}{$r_{t} = 1$}{$p_{t} = 1$}
	\AddRow{2019-06-24-21-30-17-452}{5}{4o}{$r_{t} = 1$}{$p_{t} = 1$}
	
    \node[textbox, above of=1o] {\small $\hat{s}_{t}$};
	\node[textbox, above of=1a] {\small $\hat{s}_{t+1}$};
	\node[textbox, above of=1bc, node distance=18] {\small BicycleGAN};
	\node[textbox, above of=1pc] {\small Pix2Pix};
	\node[textbox, above of=1vc] {\small VAE};
	
	\node[right of=cleft, node distance=253] (cright) {};

	\AddRow{2019-05-07-20-11-38-717}{6}{cright}{$r_{t} = 1$}{$p_{t} = 3$}
	\AddRow{2019-07-01-11-28-34-816}{7}{6o}{$r_{t} = 1$}{$p_{t} = 1$}
	\AddRow{2019-07-01-14-05-53-150}{8}{7o}{$r_{t} = 1$}{$p_{t} = 1$}
	\AddRow{2018-12-11-09-54-26-687}{9}{8o}{$r_{t} = 0.8$}{$p_{t} = 4$}
	\AddRow{2019-07-01-11-10-31-450}{10}{9o}{$r_{t} = 0$}{$p_{t} = 1$}
	
    \node[textbox, above of=6o] {\small $\hat{s}_{t}$};
	\node[textbox, above of=6a] {\small $\hat{s}_{t+1}$};
	\node[textbox, above of=6bc, node distance=18] {\small BicycleGAN};
	\node[textbox, above of=6pc] {\small Pix2Pix};
	\node[textbox, above of=6vc] {\small VAE};
\end{tikzpicture}
	    \caption{Comparison of predictions by the BicycleGAN, Pix2Pix (without uncertainty) and VAE model. }
		\label{subfig:example-small-predictions-comparison}
	\end{subfigure}
	
	\begin{subfigure}[t]{\linewidth}
		\centering
\begin{tikzpicture}
    \tikzstyle{pred} = [node distance=\graphicssize]
    \tikzstyle{textbox} = [node distance=19]
    
    \node[] (cleft) {};

	\AddColorRow{2020-06-25-21-02-27-805}{1}{cleft}{$r_{t} = 1$}{$p_{t} = 1$}
	\AddColorRow{2020-07-28-19-49-21-994}{2}{1colorbefore}{$r_{t} = 1$}{$p_{t} = 0$}
	\AddColorRow{2020-07-28-01-40-42-547}{3}{2colorbefore}{$r_{t} = 1$}{$p_{t} = 0$}
	\AddColorRow{2020-07-29-11-15-13-630}{4}{3colorbefore}{$r_{t} = 1$}{$p_{t} = 2$}
	\AddColorRow{2020-07-27-21-41-02-304}{5}{4colorbefore}{$r_{t} = 1$}{$p_{t} = 1$}
	
    \node[textbox, above of=1imagecenter] {\small $\hat{s}_{t}$};
	\node[textbox, above of=1colorcenter] {\small $\hat{s}_{t+1}$};
	\node[textbox, above of=1bc] {\small BicycleGAN};
	\node[textbox, above of=1predcolorcenter] {\small Uncertainty};

	\node[right of=cleft, node distance=253] (cright) {};

	\AddColorRow{2020-07-02-14-18-59-404}{6}{cright}{$r_{t} = 1$}{$p_{t} = 3$}
	\AddColorRow{2020-07-27-23-52-19-678}{7}{6colorbefore}{$r_{t} = 1$}{$p_{t} = 3$}
	\AddColorRow{2020-07-29-14-42-23-427}{8}{7colorbefore}{$r_{t} = 1$}{$p_{t} = 1$}
	\AddColorRow{2020-07-08-19-03-07-681}{9}{8colorbefore}{$r_{t} = 1$}{$p_{t} = 1$}
	\AddColorRow{2020-07-29-11-47-18-573}{10}{9colorbefore}{$r_{t} = 1$}{$p_{t} = 2$}
	
    \node[textbox, above of=6imagecenter] {\small $\hat{s}_{t}$};
	\node[textbox, above of=6colorcenter] {\small $\hat{s}_{t+1}$};
	\node[textbox, above of=6bc] {\small BicycleGAN};
	\node[textbox, above of=6predcolorcenter] {\small Uncertainty};
\end{tikzpicture}
        \caption{Predictions by the BicycleGAN, for both RGB (not used for downstream manipulation) and depth images.}
		\label{subfig:example-small-predictions-rgbd}
	\end{subfigure}
	
	\caption{Example predictions given an image window $\hat{s}_t$ before a manipulation action of type~$p_t$ and reward~$r_t$. $\hat{s}_{t+1}$ corresponds to the ground truth. The normalized pixel-wise uncertainties are shown from low (blue) to high (red). Action types $p_t \in \{ 0, 1, 2, 3 \}$ are grasps of different pre-shaped gripper width; $p_t \in \{ 4, 5 \}$ are shifting motions.}
	\label{fig:example-small-predictions}
\end{figure*}

Following, we assume the independence of each state uncertainties $\hat{\sigma}_{s,t}$ over time $t$. Due to the linearity of the variance, we define the cumulative uncertainty
\begin{align}
    \Sigma_{t}^2 &= \sum_{i=0}^{t} \sigma_{s; i}^2
\end{align}
as the accumulated uncertainty of the state. Furthermore, we denote $t$ as the number of action steps after an image measurement at $t=0$. At $t=0$, we consider the sensor to be perfect and reset the cumulative uncertainty to zero. Moreover, we adapt the policy $\pi$ to select the action $a_t$ maximizing the lower confidence bound of the estimated reward $\psi_t$ given by
\begin{align}
    a_{t}^* &= \argmax_a \left( M(s_t) - \alpha \left\lvert \frac{\partial M(s_t)}{\partial s} \right\rvert \Sigma_{t} \right)
    \label{eq:lower-confidence-bound}
\end{align}
using the cumulative uncertainty $\Sigma_t$ in contrast to (\ref{eq:propagation-of-uncertainty}). Let $\alpha > 0$ denote a parameter for weighing the estimated reward of actions against their confidence. If the uncertainty rises above a threshold $\sigma_{\text{image}}^2$, a new image is taken. If the lower bound of the estimated reward falls below a threshold, the bin is assumed to be empty. 
Furthermore, given both a policy $\pi$ and a transition model $T$, states and actions can be predicted multiple steps ahead by iteratively applying $\pi$ and $T$ alternately. Fig.~\ref{fig:example-iterative-predictions} shows the computational flow of a single prediction step, allowing to plan ahead and optimize for multiple-step criteria. As we don't focus on planning algorithms itself, we implemented a simple breadth-first tree search. To allow diverse tree paths, we adapt (\ref{eq:lower-confidence-bound}) to sample from the $N \approx 5$ highest actions uniform randomly. This way, we can predict the overall tree ahead of time and return the action path maximizing the criteria during inference.

\section{EXPERIMENTAL RESULTS}

Our experiments are performed on a Franka Panda robot arm using the default gripper and a flange-mounted Ensenso N10 or Framos D435e stereo camera (Fig.~\ref{fig:front-page}). We designed and printed custom gripper jaws and attached household anti-slip silicone roll to the robot's fingertips. Otherwise, the friction of the jaws would not be sufficient for shifting objects. The system uses an Intel Core i7-8700K processor and a NVIDIA GeForce GTX 1070 Ti for computing. We use the \textit{Frankx} library for robot control \cite{berscheid2021jerk}. Supplementary videos, results, as well as the source code are published at \url{https://pantor.github.io/learning-transition-for-manipulation}.

\subsection{Image Prediction}

During around \SI{180}{h} of real-world data collection, the robot attempted \num{40000}~grasps and \num{2400}~shifts. The robot used two bins with a variety of objects for training. After a successful grasp, the robot moves the objects to the other bin, however returns after filing to take an image \textit{after} the manipulation. After \num{4000} random actions, the robot used the $\varepsilon$-greedy exploration strategy for learning manipulation. $\varepsilon$ was reduced continuously and reached zero at \num{30000} actions, leading to an average reward of $\overline{r}\,=\,\num{0.55}$. The manipulation \ac{NN} was trained around every \num{500} actions. In the end, we train the BicycleGAN architecture on the complete dataset of state-action-state-tuples $(s_{t=0}, p_{t=0}, r_{t=0}, s_{t=1})$.
To stabilize the training of the \acp{GAN}, we smooth the labels and add random noise to the discriminator input. In comparison to the original implementation, we use a training batch size of \num{32}. Otherwise, we keep the training process of the BicycleGAN architecture. Fig.~\ref{fig:example-small-predictions} shows example predictions given the manipulation primitive type $p_t$ and the reward $r_t$. We include the ground truth as well as results of the Pix2Pix architecture (without uncertainty estimation) and a \ac{VAE} for visual comparison. Predicting both a state of $64 \times 64$ pixels with corresponding uncertainties using $N=20$ samples needs around \SI{60}{ms}.

While our transition model is able to generate realistic images, we find two common failure modes: First, low-level failures like large-scale deviations in the background or blurred edges. Second, wrong high-level features like cropped, blurred, or visual fused objects. The latter case occurs in particular for either shifting actions or occluded layers beneath the grasped object. Regions of high predicted uncertainty can be found primarily near edges, in particular of regions without depth information (black), and within the manipulated objects itself. Examples of iterative predictions are shown in Fig.~\ref{fig:example-iterative-predictions} and \ref{fig:front-page}.

\subsection{Grasp Rate}

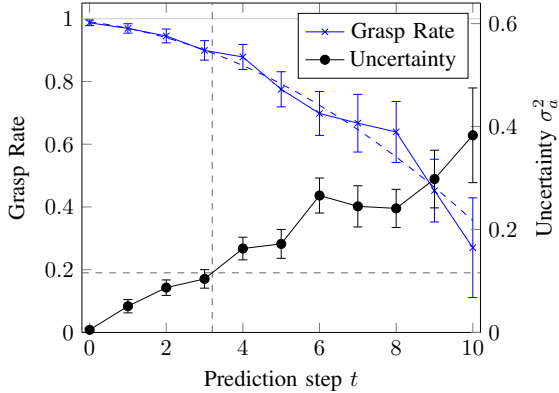
\begin{figure}[ht]
	\centering

\begin{tikzpicture}[scale=0.86]
\pgfplotsset{xmin=-0.2, xmax=10.2}

\begin{axis}[
	axis y line*=left,
	ymin=0, ymax=1.05,
	xlabel=Prediction step $t$,
	ylabel=Grasp Rate,
	legend pos=south west,
	height=190,
]
    \addplot[mark=x, blue]
        plot [error bars/.cd, y dir = both, y explicit]
        table [y=grasprate, x=step, y error=grasprate_std]{figures/grasp-rate-over-prediction-step.txt};
	\label{plot_grasprate}
	
	\addplot[mark=none, black!20] coordinates {(-0.2, 1) (10.2, 1)};
	\addplot[mark=none, dashed, black!60] coordinates {(3.2, 0) (3.2, 1.05)};
	\addplot[mark=none, dashed, black!60] coordinates {(-0.2, 0.19) (10.2, 0.19)};

	\addplot[domain=0:10, dashed, color=blue]{-0.00476*x^2 - 0.01575*x + 0.99};
\end{axis}

\begin{axis}[
	axis y line*=right,
	axis x line=none,
	ymin=0.0, ymax=0.64,
	ylabel=Uncertainty $\sigma_a^2$,
	legend pos=north east,
	height=190,
]
	\addlegendimage{/pgfplots/refstyle=plot_grasprate}\addlegendentry{Grasp Rate}
	\addplot[mark=*, black]
	    plot [error bars/.cd, y dir = both, y explicit]
	    table [y=uncertainty, x=step, y error=uncertainty_std]{figures/grasp-rate-over-prediction-step.txt};
	\addlegendentry{Uncertainty}
	
\end{axis}

\end{tikzpicture}

	\caption{Grasp rate and uncertainty of the estimated reward $\sigma_a^2$ depending on the number of prediction steps after an initial image measurement at $t=0$. Based on \num{1080} grasp attempts in the defined bin picking scenario.}
	\label{fig:grasp-predictions}
\end{figure}

We define the grasp rate as the percentage of successful grasps for grasping $15$ objects out of a bin with $25$ objects without replacement. For a bin picking scenario with multiple object types, Fig.~\ref{fig:grasp-predictions} shows the dependency between the grasp rate and the prediction step $t$, equivalent to the number of actions since the last image measurement.
The grasp rate decreases from \SI{98.7}{\%} at $t=0$ to around \SI{90}{\%} for $t=3$. Then, the average uncertainty of the estimated reward $\sigma_a^2$ increases to around $\num{0.1}$, resulting in a significant term in (\ref{eq:lower-confidence-bound}).

In order to minimize the sum of lost time due to both grasping mistakes and image acquisitions, we estimate an optimal uncertainty threshold $\sigma_{\text{image}}^2$ given a rough estimate of the execution times of the robot. We identified the durations $t_G = \SI{6.5}{s}$ of the grasp and $t_I = \SI{2.5}{s}$ of the image measurement and related motions. The \ac{PPH} are then given by
\begin{align}
    \text{PPH} &= \frac{\text{Grasp Rate}}{t_G + \frac{1}{t} t_I} \approx \frac{\text{-}0.00354\,t^2 - 0.0228\,t + 0.987}{6.5 + 2.5\,t^{-1}}
\end{align}
with the optimal prediction step $t^* \approx \num{3.2}$. The quadratic regression is plotted in Fig.~\ref{fig:grasp-predictions}. To optimize the \ac{PPH} in further experiments, we set the threshold $\sigma_{\text{image}}^2$ for taking a new image to the corresponding uncertainty $\sigma_a^2 \approx \num{0.12}$.

\subsection{Benchmarking Picks Per Hour}

We evaluate our robotic system against the Box and Blocks Test from the YCB benchmark suite \cite{calli_benchmarking_2015}. Within \SI{2}{min}, the robot should grasp and place as many cubes (\SI{2.5}{cm} side length) as possible out of one bin into another.
Additionally, we introduce a setting where the robot aims to grasp multiple objects at once. Based on the work of semantic grasping \cite{jang_end--end_2017}, we extend the manipulation \ac{NN} to predict whether the closed gripper distance will be larger than \SI{4}{cm}. This corresponds to grasping at least \num{2} objects; the policy $\pi$ then maximizes the lower bound of the expected number of grasped objects.
\begin{table}[ht]
	\centering
	\vspace{1.4mm}
	\caption{Results for the Box and Blocks Test, including the grasp success rate, the object count after \SI{2}{min} and the corresponding \acf{PPH}.}
	\begin{tabular}{|l|c|c|c|}
	\hline
	& \textbf{Grasp\,Rate} & \textbf{Count} & \textbf{PPH} \\
	\hline
	Heuristic \cite{calli_benchmarking_2015} & \SI{80}{\%} & \num{10.8} & \num{324} \\
	Random & \SI{13 \pm 4}{\%} & \num{2.2 \pm 0.7} & \num{66 \pm 20} \\
	\hline
	Single & \SI{97 \pm 2}{\%} & \num{12.8 \pm 0.3} & \num{384 \pm 10} \\
	Single, Prediction & \SI{94 \pm 2}{\%} & \num{16.4 \pm 0.5} & \num{492 \pm 14} \\
	Multiple & \SI{96 \pm 2}{\%} & \num{20.4 \pm 1.0} & \num{612 \pm 29} \\
	Multiple, Prediction & \SI{94 \pm 2}{\%} & \num{23.4 \pm 0.5} & \num{702 \pm 14} \\
	\hline
	\end{tabular}
	\label{tab:blocks-test-benchmark}
\end{table}
Table~\ref{tab:blocks-test-benchmark} shows the grasp rate, object count and final \ac{PPH} for each combination of grasps with/without prediction and of single/multiple objects, in comparison to the heuristic of Calli et al. \cite{calli_benchmarking_2015}. Random grasps are evaluated by sampling randomly from the entire action space. We repeated every experiment \num{5} times. Our robot achieves a peak performance of \SI{23.5 \pm 0.5}{grasps} with \SI{3.3 \pm 0.2}{images}, corresponding to \SI{702 \pm 14}{PPH}. On average, using an approach with prediction improves the \ac{PPH} by \SI{15}{\%}.

\subsection{Planning}

We demonstrate the ability to plan ahead for the task of emptying a bin with as fewest actions as possible.
%
%
Given an object configuration of three cubes in a row at the side of the bin the robot is able to grasp all objects in \num{4} steps by shifting the middle cube first. If an outer cube is shifted otherwise, the robot needs at least \num{5} steps. In this configuration, planning can reduce the average number of action steps from \num{4.77 \pm 0.2} to \num{4.14 \pm 0.13}, measured by \num{25} successful episodes with a success rate of \SI{78}{\%}. Illustrative material can be found on the project website.

\subsection{Generalization}

\begin{figure}[b]
	\centering
\newcommand{\graphicssize}{0.1\linewidth}
\newcommand\IncludeGraphic[2]{
    \includegraphics[width=\graphicssize]{figures/generalization/#1/#2}
}

\newcommand\AddColorRow[5]{
    \node[pred, below of=#3] (#2colorbefore) {\IncludeGraphic{#1}{bi_s_bef_color.png}};
	\node[pred, right of=#2colorbefore] (#2depthbefore) {\IncludeGraphic{#1}{bi_s_bef_depth.png}};
	\node (#2beforecenter) at ($(#2colorbefore)!0.5!(#2depthbefore)$) {};
	
	\node[left of=#2colorbefore, node distance=28] (#2labelcenter) {};
	\node[text width=24, align=right, above of=#2labelcenter, node distance=4] {\footnotesize #4};
	\node[text width=24, align=right, below of=#2labelcenter, node distance=6] {\footnotesize #5};
	
	\node[pred, right of=#2depthbefore, node distance=30] (#2colorafter) {\IncludeGraphic{#1}{bi_s_aft_color.png}};
	\node[pred, right of=#2colorafter] (#2depthafter) {\IncludeGraphic{#1}{bi_s_aft_depth.png}};
	\node (#2aftercenter) at ($(#2colorafter)!0.5!(#2depthafter)$) {};

	\node[pred, right of=#2depthafter, node distance=30] (#2colorpred) {\IncludeGraphic{#1}{bi_result_color.png}};
	\node[pred, right of=#2colorpred] (#2depthpred) {\IncludeGraphic{#1}{bi_result_depth.png}};
	\node (#2predcenter) at ($(#2colorpred)!0.5!(#2depthpred)$) {};
	
	\node[pred, right of=#2depthpred, node distance=30] (#2colorunc) {\IncludeGraphic{#1}{bi_unc_color.png}};
	\node[pred, right of=#2colorunc] (#2depthunc) {\IncludeGraphic{#1}{bi_unc_depth.png}};
	\node (#2unccenter) at ($(#2colorunc)!0.5!(#2depthunc)$) {};
}

\begin{tikzpicture}
    \tikzstyle{pred} = [node distance=\graphicssize]
    \tikzstyle{textbox} = [node distance=19]
    
    \node[] (cleft) {};

	\AddColorRow{2020-08-13-13-28-49-073}{1}{cleft}{$r_{t} = 1$}{$p_{t} = 1$}
	\AddColorRow{2020-08-13-13-31-50-939}{2}{1colorbefore}{$r_{t} = 1$}{$p_{t} = 2$}
	\AddColorRow{2020-08-13-13-31-01-577}{3}{2colorbefore}{$r_{t} = 1$}{$p_{t} = 2$}
	\AddColorRow{2020-08-13-13-32-30-506}{4}{3colorbefore}{$r_{t} = 1$}{$p_{t} = 1$}
	\AddColorRow{2020-08-13-13-33-21-639}{5}{4colorbefore}{$r_{t} = 1$}{$p_{t} = 1$}
	\AddColorRow{2020-08-13-13-43-17-473}{7}{5colorbefore}{$r_{t} = 1$}{$p_{t} = 0$}
	\AddColorRow{2020-08-13-13-38-30-376}{8}{7colorbefore}{$r_{t} = 1$}{$p_{t} = 1$}
	
    \node[textbox, above of=1beforecenter] {\small $\hat{s}_{t}$};
	\node[textbox, above of=1aftercenter] {\small $\hat{s}_{t+1}$};
	\node[textbox, above of=1predcenter] {\small BicycleGAN};
	\node[textbox, above of=1unccenter] {\small Uncertainty};
\end{tikzpicture}
	
	\caption{Example predictions for novel (unknown) objects. The uncertainties are shown from low (blue) to high (red).}
	\label{fig:generalization-predictions}
\end{figure}
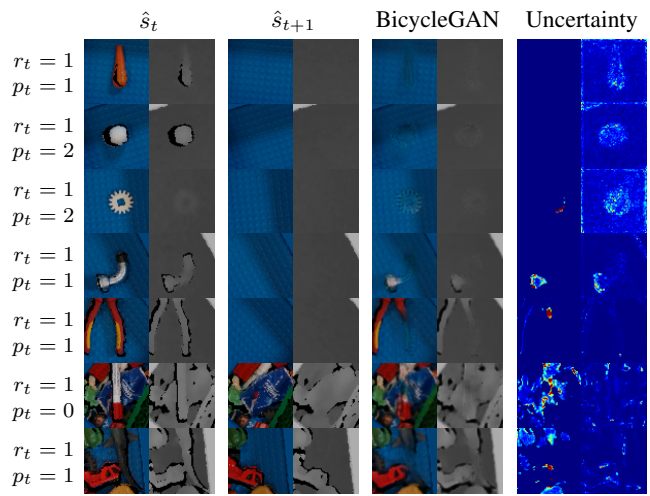

We qualitatively evaluate the system for objects that were never seen during training (Fig.~\ref{fig:generalization-predictions}).
We find that our system is able to generalize well to compact objects, naturally depending on the similarity to trained objects. From the perspective of instance segmentation, this problem is very challenging for unknown objects, as the system needs to estimate the connection of objects. For example, the pincers (5th image) are only connected outside out of the image. For long and irregular objects, the model often splits single objects into multiple parts. However, it often captures the boundaries for other objects well, e.g. like the red object in the bottom row.

\section{DISCUSSION AND OUTLOOK}

We presented an approach to learn a generative transition model for robotic manipulation, for both the next visual state and its pixel-wise uncertainty. This way, the robot is able to plan sequences of single-step manipulation primitives, which were learned data-efficiently without sparse rewards. Furthermore, we demonstrated two more novel skills of our robot: By skipping image measurements, it was able to increase the \ac{PPH} by around \SI{15}{\%} in the Box and Blocks Test \cite{calli_benchmarking_2015}. Second, the robot was able to optimize regarding multiple-step criteria in a flexible way, e.g. to minimize the number of actions for emptying a bin. This becomes possible by three major contributions: First, we simplify the transition model by conditioning it on the already learned reward estimation as well as using the spatial invariance of the state space. Second, we propose a model architecture generating samples from a multi-modal distribution, allowing to estimate a prediction uncertainty. Third, the latter was propagated efficiently towards the estimated reward. \\

Our transition model works in a direct image-to-image setting similar to \cite{boots_2014_learning, finn2019foresight, ebert2018visual}. In comparison, our model additionally estimates a pixel-wise uncertainty measure and is applied to more difficult, obstacle-rich environments. From a broader perspective, we see several advantages in our approach: In comparison to model-based \ac{RL}, our transition model does not interact with the single-step policy during training and extends the robotic system \textit{optionally}. This allows to condition the transition model on the estimated reward and reuse the manipulation model learned with model-free \ac{RL}. Oftentimes, a transition model is used for model-based policy training to reduce sample complexity \cite{hafner2019learning}. However, we find that this is not useful for grasping: The following states depend strongly on the grasp success, which can be measured easily by the gripper and learned explicitly by a manipulation model. \\

Still, our approach allows for flexible optimization criteria and - even more important for real-world applications - off-policy \ac{RL} of time-dependent manipulation sequences. In this regard, we learned simple manipulation primitives similar to \cite{zeng_learning_2018}, but were able to combine multiple primitives to more complex action sequences. Although model errors are inevitable for robotic manipulation \cite{finn2019foresight, ebert2018visual}, our uncertainty-aware policy does \textit{actively} avoid these model limitations.
However, while our transition model is able to generate realistic images, we find that it fails to capture high-level features of the stochastic physics. Instead, it usually learns the simplest solution and provides low-level pixel uncertainties. \\

Eventually, our work leaves room for several improvements. So far, we have ignored the distribution shift between real and predicted images, and in particular its effect on the manipulation model. Despite the principle ability to generalize to unknown objects, we focused our bin picking evaluation on a few object types. We plan to further investigate the limits of generalization in manipulation experiments. 
Regarding uncertainty estimation, eliminating our sample-based approach could accelerate the algorithm significantly. Moreover, we plan to integrate additional uncertainties, particularly of the manipulation \ac{NN}, into the policy.

\bibliographystyle{IEEEtran}
\bibliography{./library}

\end{document}